\documentclass[letterpaper, 10 pt, conference]{ieeeconf}  %

\IEEEoverridecommandlockouts
\usepackage[pdftex]{graphicx}
\graphicspath{{figures/}}
\usepackage{hyperref}
\hypersetup{
    colorlinks=true,
    linkcolor=black,
    citecolor=black,
    filecolor=black,
    urlcolor=black,
}
\usepackage[T1]{fontenc}
\usepackage[utf8]{inputenc}
\usepackage{csquotes}
\usepackage[english]{babel}
\usepackage[export]{adjustbox}
\usepackage[font=small]{caption}
\usepackage{subcaption}
\usepackage[colorinlistoftodos]{todonotes}
\usepackage{lipsum}
\usepackage[ruled,linesnumbered]{algorithm2e}
    \SetKwComment{Comment}{$\triangleright$\ }{}
    \SetCommentSty{lmodern} %
\usepackage[per-mode=fraction]{siunitx}
\usepackage{tikz}
\usepackage{bm}
\usepackage{colortbl}
\usepackage{multirow}
\usepackage{amsmath,amstext,amssymb}
\usepackage{mathtools}
\usepackage{placeins}%
\usepackage{carshapes}
\usepackage{mathtools}
\usepackage{mathrsfs}
\usepackage[capitalize]{cleveref}
\usepackage{balance}
\usepackage{flushend}

\usepackage[style=ieee,
            doi=false,
            url=false,
            mincitenames=1,
            maxcitenames=1,
            minbibnames=6,
            maxbibnames=6,
            backend=biber]{biblatex}  %
\title{%
Safety Reinforced Model Predictive Control (SRMPC): Improving MPC with Reinforcement Learning for Motion Planning in Autonomous Driving
}

\author{
    Johannes Fischer$^{\,\ast}$$^{\,1}$,
    Marlon Steiner$^{\,\ast}$$^{\,1}$,
    {\"O}mer {\c S}ahin Ta{\c s}$^{\,2}$ and
    Christoph Stiller$^{\,1}$%
    \thanks{
        $^{\ast}$Authors have equal contribution.
    }%
    \thanks{
        $^{1}$ Johannes Fischer, Marlon Steiner, Christoph Stiller are with the Institute of Measurement and Control Systems, Karlsruhe Institute of Technology (KIT),
        Karlsruhe, Germany.
        {\tt\small \{johannes.fischer, marlon.steiner, stiller\}@kit.edu}
    }%
    \thanks{
        $^{2}$ {\"O}mer {\c S}ahin Ta{\c s} is with FZI Research Center for Information Technology,
        Karlsruhe, Germany.
        {\tt\small tas@fzi.de}
    }%
}

\DeclareMathOperator*{\argmax}{arg\,max}

\newcommand{\norm}[1]{\left\lVert#1\right\rVert}

\definecolor{green1}{rgb}{0.4660, 0.6740, 0.1880}
\definecolor{blue1}{rgb}{0, 0.4470, 0.7410}
\definecolor{red1}{rgb}{0.6350, 0.0780, 0.1840}

\AtBeginBibliography{\small}

\usepackage{textcomp}
\usepackage[absolute,showboxes]{textpos}
\usepackage{setspace}

\TPMargin{5pt}

\newcommand{\acceptancenotice}[2]{  %
    \begin{textblock}{0.82}(0.09,0.935)
        \setstretch{0.65} %
        \noindent{\footnotesize{\copyright #1 IEEE.
        Personal use of this material is permitted.
        Permission from IEEE must be obtained for all other uses, in any current or future media,
        including reprinting/republishing this material for advertising or promotional purposes,
        creating new collective works, for resale or redistribution to servers or lists,
        or reuse of any copyrighted component of this work in other works.
        DOI: \url{#2}

        \vspace{2mm}
        \noindent
        In Proceedings of the IEEE International Conference on Intelligent Transportation Systems (ITSC) -- Bilbao, \textsc{Spain}, 24-28 September 2023.
        }}%
    \end{textblock}
}

\begin{document}
\maketitle

\acceptancenotice{2023}{https://doi.org/10.1109/ITSC57777.2023.10422605}

\thispagestyle{empty}
\pagestyle{empty}

\begin{abstract}

Model predictive control (MPC) is widely used for motion planning, particularly in autonomous driving. 
Real-time capability of the planner requires utilizing convex approximation of optimal control problems (OCPs) for the planner. 
However, such approximations confine the solution to a subspace, which might not contain the global optimum.
To address this, we propose using safe reinforcement learning (SRL) to obtain a new and safe reference trajectory within MPC. 
By employing a learning-based approach, the MPC can explore solutions beyond the close neighborhood of the previous one, potentially finding global optima. 
We incorporate constrained reinforcement learning (CRL) to ensure safety in automated driving, using a handcrafted energy function-based safety index as the constraint objective to model safe and unsafe regions. 
Our approach utilizes a state-dependent Lagrangian multiplier, learned concurrently with the safe policy, to solve the CRL problem. 
Through experimentation in a highway scenario, we demonstrate the superiority of our approach over both MPC and SRL in terms of safety and performance measures.

\end{abstract}

\vspace{0.5em}

\section{Introduction}
\label{introduction}

Autonomous vehicles have the potential to fundamentally change traffic by reducing risky driving behavior, congestion, carbon dioxide emissions, and transportation costs, as well as improving road safety \cite*{Li.2022}.
An important area in the field of autonomous driving is motion planning, which aims to generate optimal trajectories in a given state that at the same time do not pose a threat to road traffic.
Motion planning is often performed by model predictive control (MPC) \cite{Mousavi.2013, Inghilterra.2018, Yu.2020, Wang.2022, tas2022motion}.
For real-time applications, linear time-varying MPC (LTV-MPC) is often preferred because of the ability to faster solution calculation \cite{Lima.2017,Jain.2019}. 
Though, LTV-MPC only performs a local optimization which leads to the fact that only a locally optimal solution can be found.

Recent studies also focus on motion planning and control with reinforcement learning (RL).
In RL applications it is a huge challenge to address constraint satisfaction, especially in safety critical applications like autonomous driving.
There is a lot of research trying to achieve safety in RL, which is called safe RL (SRL) \cite{garcia.2015,Ray.2019,Ma.CRL.2021,Qin.2021,Ma.2022,fischerSamplingbasedInverseReinforcement2021,kamranMinimizingSafetyInterference2021,kamranModernPerspectiveSafe2022}.
Most of the latest approaches are not tested on driving tasks yet, and they often require further research. 
However, full autonomous driving is still confronted with the task to make feasible and especially safe decisions and planning in complex and uncertain dynamic traffic scenarios.

In order to address the existing issues, we present a novel approach which combines both methods, MPC and RL, to improve the motion planning in terms of safety and performance. 
This leads to a novel method that combines the mathematical optimization approach with a learning-based approach.
Our main contributions are:
\begin{itemize}
    \item The application of a SRL approach with state dependent Lagrange multiplier and safety index for autonomous driving tasks.
    \item Utilization of SRL for switching between local optima and ideally finding the global optimum, which serves as the reference for the locally optimizing MPC.
    \item Experimental results on a highway scenario, demonstrating that the proposed method achieves better results in terms of safety and performance than SRL and LTV-MPC.
\end{itemize}

\section{Related Work}
\label{sec:related_work}

To the best of our knowledge there are no other approaches, which use RL to improve the motion planning of an MPC controller by generating a reference trajectory.
There are several approaches of so called \textit{learning-based MPC}, which use machine learning to learn dynamic models or parameters of the MPC controller \cite{Williams.2017,Nghiem.2019,Aswani.2013,Ostafew.2016,Koller.2019}.
RL in combination with MPC is often used to imitate the control strategy of the MPC by \textit{end-to-end learning} \cite{Amos.2018,Chen.2020,Drgona.2021}.
However, approaches where the MPC benefits from the learned value function of the RL algorithm can rarely be found in the literature.
The authors in \cite{Arroyo.2022} integrate the RL value function into the objective function of the MPC.
But they solved the optimal control problem (OCP) in a combinatorial way and not analytically.
This makes the proposed approach only applicable for problems with a small and discrete action space.
Another approach which is closely related to ours is presented in \cite{Greatwood.2019}.
In this paper the authors use RL to define a terminal area for the MPC, so that the controlled vehicle can be guided through an unknown environment.
In contrast to \cite{Greatwood.2019}, we do not define a terminal area but use RL to generate a whole trajectory as a reference respectively initial solution for the OCP.
Furthermore, the learned trajectory in our work is not forced to be tracked by the MPC but can shift the optimization problem from one locally optimal solution to another.

\section{Fundamentals}
\label{sec:background}

In this section, we formalize the definition of the MPC and of the RL problem as a constrained Markov decision process (CMDP) and present the algorithms that are used to solve it.

\subsection{MPC and Linear Time-Varying MPC}
MPC is a control strategy that predicts future system behavior and determines optimal control actions $\bm{u}$ based on the modeled system dynamics.
It computes control actions for a finite time horizon by solving an OCP, but only applying the first action to the system.
The process is repeated at every time step, taking into account new measurements and updated predictions for the state $\bm{x}$.
Due to often nonlinear constraints in the OCP within an autonomous driving task, a nonlinear MPC (NMPC) for the control of such a system is required.
The NMPC solves an optimization problem
\begin{subequations}\label{OCP_tracking}
    \begin{align}
        \underset{\bm{u}_{1:N},\bm{x}_{1:N+1}}{\text{min}} \quad & J(\bm{u}_{1:N},\bm{x}_{1:N+1}) \label{OCP_a}                                                                                          \\
        \text{s.t.} \quad                                        & \bm{x}_{k+1} = \mathbf{f}(\bm{x}_k,\bm{u}_k),\ k=1,\ldots,N, \label{OCP_b}                                                            \\
                                                                 & \bm{x}_1=\hat{\bm{x}}, \label{OCP_c}                                                                                                  \\
                                                                 & \underline{\mathbf{g}_{\text{col}}} \leq \mathbf{g}_{\text{col}}(\bm{x}_k,\bm{u}_k) \leq \overline{\mathbf{g}_{\text{col}}},\nonumber \\
                                                                 & \phantom{---------} k=2,\ldots,N\!+\!1, \label{OCP_d}                                                                                 \\
                                                                 & \underline{\bm{u}} \leq \bm{u}_k \leq \overline{\bm{u}},\ k=1,\ldots,N, \label{OCP_e}                                                 \\
                                                                 & \underline{\bm{x}} \leq \bm{x}_k \leq \overline{\bm{x}},\ k=1,\ldots,N\!+\!1, \label{OCP_f}
    \end{align}
\end{subequations}
in every time step specified by the objective function \eqref{OCP_a}, the system dynamics \eqref{OCP_b}, the initial state condition \eqref{OCP_c}, collision avoidance constraints \eqref{OCP_d} denoted by general collision functions $\mathbf{g}_{\mathrm{col}}$ and box constraints for the control and state variables \eqref{OCP_e}--\eqref{OCP_f}.
In the OCP \eqref{OCP_tracking} the variable $N$ represents the prediction horizon length.

Applying NMPC to an on-board system is challenging due to real-time constraints and the computational complexity of the optimization problem.
LTV-MPC addresses this challenge by incorporating time-varying dynamics through the linearization of the optimization problem constraints at every time step.
Together with a quadratic objective function to optimize, the problem becomes convex and a quadratic programming (QP) solver can be applied to solve it.
However, this leads to the fact that only a locally optimal solution is calculated which depends on the trajectory used for linearization.

\subsection{Constrained Markov Decision Process}
Analogous to a Markov decision process (MDP) \cite{Sutton.2018}, a CMDP \cite{Altman.1999} is a model that describes a problem of sequential decision-making between an agent and its environment.
The difference is that the CMDP allows a clear separation of reward and safety signals.
A CMDP can be defined as a tuple $(\mathcal{S},\mathcal{A},\mathcal{P}_0,r,c,d,\gamma)$, where $\mathcal{S}$ is the state space, $\mathcal{A}$ is the action space, $\mathcal{P}_0\colon\mathcal{S}\!\times\!\mathcal{A}\!\times\!\mathcal{S}\!\to\![0,1]$ is the transition probability kernel indicating the probability of state $s'$ after taking action $a$ in state $s$, $r\colon\mathcal{S}\times\mathcal{A}\to\mathbb{R}$ is the reward function, $c\colon\mathcal{S}\times\mathcal{A}\to\mathbb{R}$ is the cost function, $d$ is the safety threshold and $\gamma$ is the discount factor.

The goal in a CMDP is to find a policy $\pi$ that maximizes the accumulated discounted reward
\begin{equation}
    \underset{\pi}{\text{max}} \ J_r(\pi) = \underset{\tau\sim\pi}{\mathbb{E}}\left[ \sum_{t=0}^{\infty}\gamma^t r(s_t,a_t) \right] \label{CMDP_a}
\end{equation}
while keeping the accumulated discounted cost
\begin{equation}
    J_{c}(\pi) = \underset{\tau\sim\pi}{\mathbb{E}}\left[ \sum_{t=0}^{\infty}\gamma_c^t c(s_t,a_t) \right] \leq d \label{CMDP_b}
\end{equation}
bounded to $d$ \cite{CPO}.
Here $\tau$ denotes a trajectory ${\tau=(s_0,a_0,s_1,\ldots)}$ distributed according to the policy $\pi$.
Equation \eqref{CMDP_b} represents the constraint of the optimization problem.
It is possible for the cost function to represent either a physically based function or an indicator function with discrete values for safety.
In this context costs occur when the constraint is violated.

The set of feasible stationary policies that satisfy the constraint of the defined CMDP is then given by
\begin{equation} \label{Pi_feasible}
    \Pi_c \doteq \left\{ \pi\in\Pi\colon J_c(\pi) \leq d \right\}.
\end{equation}

\subsection{Constrained Reinforcement Learning}
Constrained RL (CRL) is a variant of classic RL which considers constraints in the MDP and can be classified as an approach for safe RL (SRL).
The most commonly used CRL algorithm is the Lagrangian approach where the optimization criterion \eqref{CMDP_a} is extended by the cost constraint \eqref{CMDP_b} \cite{CVPO.2022}.
Examples for baseline CRL Lagrangian algorithms are PPO-Lagrangian and TRPO-Lagrangian \cite{Ray.2019}.
For detailed information about the classification of SRL approaches we refer the reader to \cite{garcia.2015}.

The goal of CRL is to learn a feasible optimal policy $\pi^\ast$ regarding \eqref{Pi_feasible} that maximizes the objective $J_r$:

\begin{equation}\label{rl_goal}
    \pi^\ast = \argmax_{\pi\in\Pi_c} \, J_r(\pi).
\end{equation}

Within the Lagrangian approach for solving CMDPs, the objective function becomes a $\mathrm{min}$-$\mathrm{max}$ optimization problem
\begin{equation}\label{lagrangian}
    \underset{\pi}{\mathrm{min}}\,\underset{\lambda\geq0}{\mathrm{max}}\ \mathcal{L}(\pi,\lambda)=-J_r(\pi)+\lambda (J_c(\pi)-d)
\end{equation}
to balance between reward and cost.
The optimal solution to \eqref{lagrangian} is given by its saddle point.

\subsection{Energy Function based Safety}
\label{subsec:si}
In control theory a control strategy is considered as safe when unsafe regions of the state space are avoided and at the same time the safe region is forward invariant.
The safety of a control can be evaluated by a safety certificate in form of a scalar energy function $\phi\colon\mathbb{R}^n\to\mathbb{R}$.
A function like this is also called safety index (SI).
In general those functions take on positive values to indicate unsafe states ($\phi > 0$) and take on negative values to indicate safe states ($\phi \leq 0$) \cite{Wei.2019}.

To achieve safety, energy dissipation is required so that $\dot{\phi} < 0$ holds if the state is unsafe.
According to this assumption a safe control follows the condition
\begin{equation}\label{safe-action-constr}
    \phi(s') < \text{max}\{\phi(s)-\eta,0\}
\end{equation}
on the energy of the SI.
In this context $\eta$ is a small slack variable to force the energy to decrease if the current state $s$ is unsafe.
As soon as the state $s$ gets safe the next state $s'$ has to be safe, too.
Safe states can be represented as a subset of all system states $\mathcal{S}$ by a closed, connected safe set $\mathcal{S}_s$.
The SI must be designed such that the set $\mathcal{S}_s$ is a zero-sublevel set of $\phi\colon\mathcal{S}\to\mathbb{R}$: $\mathcal{S}_s=\{s\,|\,\phi(s)\leq0\}$ \cite{Ma.2022}.

Further, the following restrictions for the design of a feasible SI must hold:

\begin{enumerate}
    \item The time derivative of the SI can be influenced by the control variables $\bm{u}$ of the system: $\frac{\partial\dot{\phi }}{\partial\bm{u}}\neq0$
    \item and the reachable set under $\phi$ is a subset of the safe set: $X(\phi)\subseteq\mathcal{X}_S$ \cite{Noren.2021}.
\end{enumerate}

\section{Method}
\label{sec:method}
In this section, we formulate the used MPC and RL methods independently to combine them afterwards into the proposed SRMPC approach.

\subsection{MPC for Automated Driving}
Subsequently, we define the relevant terms of the OCP \eqref{OCP_tracking}.
The used MPC scheme is state of the art, hence we only give a brief overview.

\subsubsection{Objective Function}
We use quadratic forms to define the objective function \eqref{OCP_a} to be optimized.
In this way, we obtain a convex function, which is a central property regarding the choice of the solution method.
The objective function is defined such that a reference trajectory $\bm{x}^\mathrm{ref}$ is tracked and the control variables and their derivations are minimized:
\begin{align}\label{OCP_objective}
    J = & \sum\limits_{k=1}^{N+1} \norm{\begin{matrix}
                                                \bm{x}_k - \bm{x}_k^\text{ref} \end{matrix}}_{\mathbf{Q}}^2
    +
    \sum\limits_{k=1}^{N} \norm{\begin{matrix}
                                        \bm{u}_k \end{matrix}}_{\mathbf{R}}^2 \nonumber                     \\
        & +
    \norm{\begin{matrix}\bm{u}_1-\hat{\bm{u}}\end{matrix}}_{\mathbf{S}}^2
    +
    \sum\limits_{k=2}^{N} \norm{\begin{matrix} \bm{u}_k - \bm{u}_{k-1} \end{matrix}}_{\mathbf{S}}^2
    +
    J_{\mathrm{slack}},
\end{align}
where $\hat{\bm{u}}$ denotes the previously applied control.
The term $J_\mathrm{slack}$ in equation \eqref{OCP_objective} penalizes the solver when using slack variables.
Slack variables are applied to the collision constraints \eqref{OCP_d} to guarantee feasibility of the problem by relaxing the whole constraint.
Additional slack variables are used to define a safety region around the vehicles which is only entered by the ego-vehicle if necessary \cite{Jain.2019}.

\subsubsection{Vehicle Model}\label{subsubsec:model}
\begin{figure}[!h]
    \centering\vspace*{2mm}
    \def\svgwidth{0.9\linewidth}
\begingroup%
  \makeatletter%
  \providecommand\color[2][]{%
    \errmessage{(Inkscape) Color is used for the text in Inkscape, but the package 'color.sty' is not loaded}%
    \renewcommand\color[2][]{}%
  }%
  \providecommand\transparent[1]{%
    \errmessage{(Inkscape) Transparency is used (non-zero) for the text in Inkscape, but the package 'transparent.sty' is not loaded}%
    \renewcommand\transparent[1]{}%
  }%
  \providecommand\rotatebox[2]{#2}%
  \newcommand*\fsize{\dimexpr\f@size pt\relax}%
  \newcommand*\lineheight[1]{\fontsize{\fsize}{#1\fsize}\selectfont}%
  \ifx\svgwidth\undefined%
    \setlength{\unitlength}{329.25214392bp}%
    \ifx\svgscale\undefined%
      \relax%
    \else%
      \setlength{\unitlength}{\unitlength * \real{\svgscale}}%
    \fi%
  \else%
    \setlength{\unitlength}{\svgwidth}%
  \fi%
  \global\let\svgwidth\undefined%
  \global\let\svgscale\undefined%
  \makeatother%
  \begin{picture}(1,0.58736428)%
    \lineheight{1}%
    \setlength\tabcolsep{0pt}%
    \put(0,0){\includegraphics[width=\unitlength,page=1]{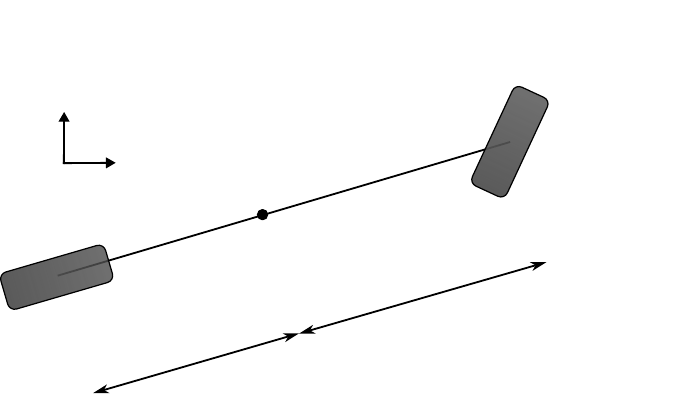}}%
    \put(0.46586167,0.36479624){\color[rgb]{0,0,0}\makebox(0,0)[lt]{\lineheight{1.25}\smash{\begin{tabular}[t]{l}$v$\end{tabular}}}}%
    \put(0.57116392,0.29130884){\color[rgb]{0,0,0}\makebox(0,0)[lt]{\lineheight{1.25}\smash{\begin{tabular}[t]{l}$\varphi$\end{tabular}}}}%
    \put(0.51657754,0.33440336){\color[rgb]{0,0,0}\makebox(0,0)[lt]{\lineheight{1.25}\smash{\begin{tabular}[t]{l}$\beta$\end{tabular}}}}%
    \put(0.25888083,0.06691497){\color[rgb]{0,0,0}\rotatebox{16.632373}{\makebox(0,0)[lt]{\lineheight{1.25}\smash{\begin{tabular}[t]{l}$l_r$\end{tabular}}}}}%
    \put(0.60555091,0.16689672){\color[rgb]{0,0,0}\rotatebox{15.502871}{\makebox(0,0)[lt]{\lineheight{1.25}\smash{\begin{tabular}[t]{l}$l_f$\end{tabular}}}}}%
    \put(0.86324895,0.46272934){\color[rgb]{0,0,0}\makebox(0,0)[lt]{\lineheight{1.25}\smash{\begin{tabular}[t]{l}$\delta$\end{tabular}}}}%
    \put(0.17182128,0.34278977){\color[rgb]{0,0,0}\makebox(0,0)[lt]{\lineheight{1.25}\smash{\begin{tabular}[t]{l}$x$\end{tabular}}}}%
    \put(0.08156913,0.43855538){\color[rgb]{0,0,0}\makebox(0,0)[lt]{\lineheight{1.25}\smash{\begin{tabular}[t]{l}$y$\end{tabular}}}}%
    \put(0.17667041,0.24555865){\color[rgb]{0,0,0}\makebox(0,0)[lt]{\lineheight{1.25}\smash{\begin{tabular}[t]{l}$F_{xr}$\end{tabular}}}}%
    \put(-0.00054689,0.22579072){\color[rgb]{0,0,0}\makebox(0,0)[lt]{\lineheight{1.25}\smash{\begin{tabular}[t]{l}$F_{yr}$\end{tabular}}}}%
    \put(0.66828154,0.43267453){\color[rgb]{0,0,0}\makebox(0,0)[lt]{\lineheight{1.25}\smash{\begin{tabular}[t]{l}$F_{yf}$\end{tabular}}}}%
    \put(0.22402967,0.20681341){\color[rgb]{0,0,0}\makebox(0,0)[lt]{\lineheight{1.25}\smash{\begin{tabular}[t]{l}$\alpha_r$\end{tabular}}}}%
    \put(0.8238716,0.56084047){\color[rgb]{0,0,0}\makebox(0,0)[lt]{\lineheight{1.25}\smash{\begin{tabular}[t]{l}$\alpha_f$\end{tabular}}}}%
    \put(0,0){\includegraphics[width=\unitlength,page=2]{bicycle.pdf}}%
    \put(0.86516443,0.51827338){\color[rgb]{0,0,0}\makebox(0,0)[lt]{\lineheight{1.25}\smash{\begin{tabular}[t]{l}$v_f$\end{tabular}}}}%
    \put(0.26758686,0.17424224){\color[rgb]{0,0,0}\makebox(0,0)[lt]{\lineheight{1.25}\smash{\begin{tabular}[t]{l}$v_r$\end{tabular}}}}%
    \put(0,0){\includegraphics[width=\unitlength,page=3]{bicycle.pdf}}%
    \put(0.32359595,0.28939352){\color[rgb]{0,0,0}\makebox(0,0)[lt]{\lineheight{1.25}\smash{\begin{tabular}[t]{l}COG\end{tabular}}}}%
  \end{picture}%
\endgroup%
     \caption{Bicycle model: The forces (red arrows) are not considered in the kinematic bicycle model and the sideslip angles are assumed to be $\alpha_{f,r}=0$.}
    \label{fig:model}
\end{figure}

For the system dynamics \eqref{OCP_b} the kinematic bicycle model \cite{kinBicycle2017} is used, see \cref{fig:model}.
The purely kinematic consideration of the vehicle dynamics can be expressed by equation \eqref{kin_1spur}:
\begin{equation}\label{kin_1spur}
    \begin{pmatrix}
        \dot{x} \\ \dot{y} \\ \dot{\varphi} \\ \dot{v}
    \end{pmatrix}
    =
    \begin{pmatrix}
        v \, \mathrm{cos}(\varphi+\beta) \\ v \, \mathrm{sin}(\varphi+\beta) \\ \frac{v}{l}\, \mathrm{tan}(\delta)\, \mathrm{cos}(\beta) \\ a
    \end{pmatrix}
\end{equation}
with the slip angle $\beta$
\begin{equation}
    \beta = \mathrm{arctan}\left(\frac{l_r}{l}\ \mathrm{tan}(\delta)\right).
\end{equation}
Thereupon, the state variables $\bm{x}$ are the global $x$- and $y$-coordinates, the heading angle $\varphi$ and the velocity $v$.
Concurrently the steering angle $\delta$ and the acceleration $a$ function are the control variables $\bm{u}$.

\subsubsection{Collision constraints}
The collision constraints \eqref{OCP_d} are the crucial part in the OCP concerning the safety.
It turned out that over-approximating the ego-vehicle by a set of circles and other traffic participants by superellipses, leads to an easy to handle mathematical form as well as good results.
The resulting collision constraint between the ego-vehicle and the $j$-th vehicle can be expressed by equation \eqref{collision-constraint}.
\begin{equation}
    \begin{bmatrix}
        \Delta x_{j} \\
        \Delta y_{j}
    \end{bmatrix}^T
    \mathbf{R}(\varphi^j)^T
    \begin{bmatrix}
        \frac{1}{a} & 0           \\
        0           & \frac{1}{b}
    \end{bmatrix}^2
    \mathbf{R}(\varphi^j)
    \begin{bmatrix}
        \Delta x_{j} \\
        \Delta y_{j}
    \end{bmatrix} > 1 \label{collision-constraint}
\end{equation}
where $\mathbf{R}$ is the rotation matrix in $\mathbb{R}^2$, $a$ and $b$ are the sum of the semi-major respectively semi-minor axis and the radius of the circle \cite{Brito.2020}.

\subsubsection{Approximate the nonlinear OCP}
We approximate the original OCP by a QP in every time step to make it applicable within the LTV-MPC method.
Therefore, we use a second order Taylor series expansion to reformulate the objective function \eqref{OCP_a} and a first order Taylor series expansion to linearize the constraints \eqref{OCP_b}--\eqref{OCP_f}.
The resulting QP has the form
\begin{equation}\label{qp}
    \begin{split}
        \underset{\mathbf{z}}{\text{min}} \quad & \frac{1}{2}\mathbf{z}^T\mathbf{H}\mathbf{z}+\mathbf{g}^T\mathbf{z} \\
        \text{s.t.} \quad                       & \mathbf{l} \leq \mathbf{A}\mathbf{z} \leq \mathbf{u}
    \end{split}
\end{equation}
with the new optimization variable $\mathbf{z}$.
Assuming the weight matrices $\mathbf{Q}$, $\mathbf{R}$ and $\mathbf{S}$ are positive semidefinite and diagonal, the QP \eqref{qp} is convex.
The Taylor expansion requires a reference trajectory $\mathbf{z}^r=[\bm{x}_{1:N+1}^r;\bm{u}_{1:N}^r]$.
To obtain such a reference, trajectory shifting \cite{Diehl.2009} is applied to the previous solution.
To clarify, this reference trajectory differs from $\bm{x}^\mathrm{ref}$ in equation \eqref{OCP_objective}.
The QP is solved using the OSQP-Solver \cite{osqp}.

\subsection{Defining the CMDP}
In this section we define the elements of the constraint optimization task, the CMDP.

\subsubsection{Observation}
\begin{figure}[!h]
    \centering
    \includegraphics[width=0.8\linewidth]{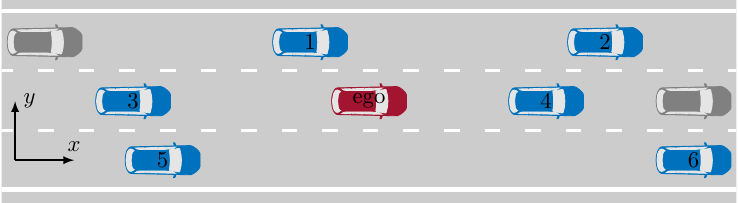}
    \caption{CMDP observation of the ego-vehicle (red): The ego-vehicle observes the vehicle in front of and the one behind itself on each lane (represented in blue). The vehicles shown in gray are not considered.}
    \label{fig:observation}
\end{figure}

:
The observation approximates the state, and functions as the input of the policy and value functions.
\cref{fig:observation} shows an exemplary scene with multiple vehicles to explain which ones are considered in the observation.

The matrices in \eqref{observation} present all considered quantities within the observation.
\begin{subequations}\label{observation}
    \begin{align}
        \begin{bmatrix}\label{observation_a}
            \Delta x^1 & \Delta y^1 & \Delta v_x^1 & \Delta v_y^1 \\
            \vdots     & \vdots     & \vdots       & \vdots       \\
            \Delta x^6 & \Delta y^6 & \Delta v_x^6 & \Delta v_y^6 \\
        \end{bmatrix}\phantom{aaaaaaa} \\
        \begin{bmatrix}\label{observation_b}
            v_x^{\text{ego}} & v_y^{\text{ego}} & \varphi^{\text{ego}} & \Delta y^{\mathrm{rs,r}} & \Delta y^{\mathrm{rs,l}} & \Delta y^{\mathrm{target}}
        \end{bmatrix}
    \end{align}
\end{subequations}
Matrix \eqref{observation_a} poses the quantities for the six surrounded vehicles (see \cref{fig:observation}).
The $\Delta$ expresses that the specific quantity is measured relative to the ego-vehicle.
Information about the ego-vehicle state and relative coordinates to the left and right roadsides (rs) and the target lane can be found in Matrix \eqref{observation_b}.

\subsubsection{Reward}
In our case the reward function $r$ is based on the deviation from a longitudinal reference velocity $v_x^\mathrm{ref}$ and a reference lane $y^\mathrm{ref}$.
We define an auxiliary reward
\begin{equation}\label{reward-function}
    r'(s,a) = - g (v_x^{\mathrm{ref}} - v_x^{\mathrm{ego}})^2 - (1-g)(y^{\mathrm{ref}} - y^{\mathrm{ego}})^2
\end{equation}
In this equation $g$ is a hyperparameter that balances the relation between speed reward and reward for driving on a specific lane.
To achieve a better and more stable performance in training, the reward $r$ is obtained by exponentiating the function $r'$ and normalizing it to the interval $[0,1]$.

\subsubsection{Actions}
The same control variables as in the MPC are used as actions to control the agent (see section \ref{subsubsec:model}).
But in contrast to the MPC, the RL algorithm outputs continuous actions due to the utilization of a stochastic policy.

\subsection{Safety Index for Highway Driving}
In this section we develop the cost function defining the CMDP.
Therefore, we define SI based on the formulation from \cite{Wei.2019,Ma.CRL.2021,Ma.2022,Liu.2014}:
\begin{equation}\label{si}
    \phi(s) = (\sigma + d_{\text{min}})^{n} - d^n - k\dot{d}
\end{equation}
with the tunable parameters $\sigma$, $n$ and $k$, the minimum distance to the other vehicle $d_\mathrm{min}$, the actual distance $d$ and the velocity towards the other vehicle $\dot{d}$.
We defined $\dot{d}$ dependent to the lateral distance from the agent to another vehicle.
When both are not on the same lane, the longitudinal velocity part of $\dot{d}$ is set to zero to make the SI \eqref{si} applicable to the highway driving.

The time derivative of the chosen SI contains both action dimensions $\delta$ and $a$, so that the relative degree of the system is $r=1$.
Under the assumption that the parameters of the SI are chosen according to the safe set $\mathcal{S}_s$, the SI fulfills both requirements for a feasible SI (see section \ref{subsec:si}).
In this paper we do not formally proof the latter condition.

The cost function results from the condition for safe control \eqref{safe-action-constr}:
\begin{equation}
    c(s,a) = \mathrm{max}\Bigl\{ \phi(s') - \text{max}\{\phi(s)-\eta,0\}, -0.1 \Bigr\}
\end{equation}
where cost occurs if the condition is not fulfilled.
The outer $\mathrm{max}$-function serves the purpose to prevent the algorithm from learning to become highly conservative.

\subsection{State dependent Lagrangian Multiplier}
As proposed in \cite{Ma.FAC.2021}, we use a state dependent Lagrangian multiplier $\lambda(s)$ to handle the state dependent constraint arising from the SI.
The value of the Lagrangian multiplier indicates the safety of the agent to be in a specific state.
It gets approximated by a state value function which is learned through SI transitions by considering the \textit{complementary slackness condition}.
The condition comes from the Karush-Kuhn-Tucker (KKT) necessary conditions of optimality.
\begin{table}[!h]
    \caption{Relation between optimal multipliers $\lambda^\ast$ and safety}\label{tab:lambda}
    \centering
    \begin{tabular}{c c l}
        \hline\noalign{\vskip 1mm}
        $\lambda^\ast(s)$ & Safety                          \\[1mm]
        \hline\noalign{\vskip 1mm}
        Zero              & Safe (inactive constraint)      \\[1mm]
        Finite            & On boundary (active constraint) \\[1mm]
        Infinite          & Unsafe (infeasible constraint)  \\[1mm]
        \hline
    \end{tabular}
\end{table}
\mbox{Table \ref{tab:lambda}} shows what the value of the optimal multiplier implies for the safety.
Numerically, the multiplier is considered infinite when exceeding a threshold we defined.

\subsection{SRL Approach to Automated Driving}
We use a state value function $v_\pi(s)$ to approximate the objective $J_r(\pi)$ \eqref{CMDP_a} and another state value function $v_\pi^c(s)$ (referred to as cost value function) to approximate the associated constraint $J_c(\pi)-d$ \eqref{CMDP_b}.
Together with the state dependent Lagrangian multiplier $\lambda(s)$ the loss function to be optimized can be rewritten as
\begin{equation}\label{lagrangian2}
    \mathcal{L}(\pi,\lambda) = \mathbb{E}_s \left[ -v_\pi(s) + \lambda(s) v_\pi^c(s)  \right].
\end{equation}

To learn a policy that solves the optimization problem, the on-policy \textit{proximal policy optimization} (PPO) \cite{PPO} baseline algorithm is adapted by the new objective \eqref{lagrangian2} and extended by the cost value function $v_\pi^c(s)$ and the state value function $\lambda(s)$ for the Lagrangian multiplier.
We call the safe version of the PPO algorithm \textit{PPO-Lagrangian-Safety Index} (PPO-L-SI).

\subsection{Combining LTV-MPC and SRL}
The limitation of LTV-MPC can be attributed to the approximating aspect of the solution computation \cite{Bender.2015}.
Due to local optimization, LTV-MPC only identifies the optimal solution within the linearization range, thereby overlooking other locally optimal solutions.
The linearization in LTV-MPC is based on the shifted solution from the previous time step, resulting in the dependency of subsequent time step solutions on the previous one.
Although this approach is computationally efficient, it neglects other local optima and possibly the global optimum of the original optimization problem.

To address this limitation, an RL policy can be employed in LTV-MPC to generate a trajectory to be used for linearization.
For the optimal use of the learned trajectories, it is crucial that the RL algorithm receives the same or a very similar optimization problem as the MPC for policy learning.
Furthermore, both control algorithms should have access to the same control variables.

The trajectory generated by the policy for the planning horizon can be used to approximate the NLP as a QP.
This provides an alternative reference trajectory compared to the shifted solution trajectory from the previous MPC step, allowing the LTV-MPC to discover a new local and potentially the global optimum.
This approach also bypasses the need for a handcrafted initial trajectory for the first time step, $t=0$, of the MPC.
For example, a trajectory following the lane with constant velocity or a braking trajectory can be used as the initial trajectory.
The proposed SRMPC approach is summarized in algorithm \ref{alg:RL-MPC}.
\begin{algorithm}[]
    \caption{Pseudocode for SRMPC}\label{alg:RL-MPC}
    \KwIn{policy $\pi$, prediction horizon length $N$}
    \SetKwInOut{Output}{Output}
    \SetKwProg{try}{try}{:}{}
    \SetKwProg{catch}{catch}{:}{end}
    $k\gets0$\;
    \While{goal is not reached}{
        $\bm{x}_k\gets$ measure current state\;
        \try{calculate trajectory with $\pi$ for linearization}{
            $\bm{x}_{k\colon k+N+1}^\mathrm{r},\bm{u}_{k\colon k+N}^\mathrm{r}\gets$ use $\pi$ to simulate environment state forward and save state and control values\;
        }
        \catch{error occurred}{
            $\bm{x}_{k\colon k+N+1}^\mathrm{r},\bm{u}_{k\colon k+N}^\mathrm{r}\gets$ use shifted trajectory from previous solution
        }
        NLP $\gets$ define and discretize the time-dependent OCP \eqref{OCP_tracking}\;
        $\mathbf{H},\mathbf{g},\mathbf{A},\mathbf{l},\mathbf{u}\gets$ calculate QP matrices and vectors (see equation \eqref{qp}) from NLP with $\bm{x}_k$,\ $\bm{x}_{k\colon k+N+1}^\mathrm{r}$,\ $\bm{u}_{k\colon k+N}^\mathrm{r}$\ \text{and}\ $\bm{x}_{k\colon k+N+1}^\mathrm{ref}$\;
        $\bm{u}_{k+1}\gets$ solve QP to get the optimal control value\;
        \KwOut{$\bm{u}_{k+1}$}
        $k\gets k+1$\;
    }
\end{algorithm}

\begin{figure}
    \def\svgwidth{\linewidth}
    \begin{tikzpicture}
        \node at (0,0) [align=center] {%
\begingroup%
  \makeatletter%
  \providecommand\color[2][]{%
    \errmessage{(Inkscape) Color is used for the text in Inkscape, but the package 'color.sty' is not loaded}%
    \renewcommand\color[2][]{}%
  }%
  \providecommand\transparent[1]{%
    \errmessage{(Inkscape) Transparency is used (non-zero) for the text in Inkscape, but the package 'transparent.sty' is not loaded}%
    \renewcommand\transparent[1]{}%
  }%
  \providecommand\rotatebox[2]{#2}%
  \newcommand*\fsize{\dimexpr\f@size pt\relax}%
  \newcommand*\lineheight[1]{\fontsize{\fsize}{#1\fsize}\selectfont}%
  \ifx\svgwidth\undefined%
    \setlength{\unitlength}{460.80001831bp}%
    \ifx\svgscale\undefined%
      \relax%
    \else%
      \setlength{\unitlength}{\unitlength * \real{\svgscale}}%
    \fi%
  \else%
    \setlength{\unitlength}{\svgwidth}%
  \fi%
  \global\let\svgwidth\undefined%
  \global\let\svgscale\undefined%
  \makeatother%
  \begin{picture}(1,1.24999995)%
    \lineheight{1}%
    \setlength\tabcolsep{0pt}%
    \put(0,0){\includegraphics[width=\unitlength,page=1]{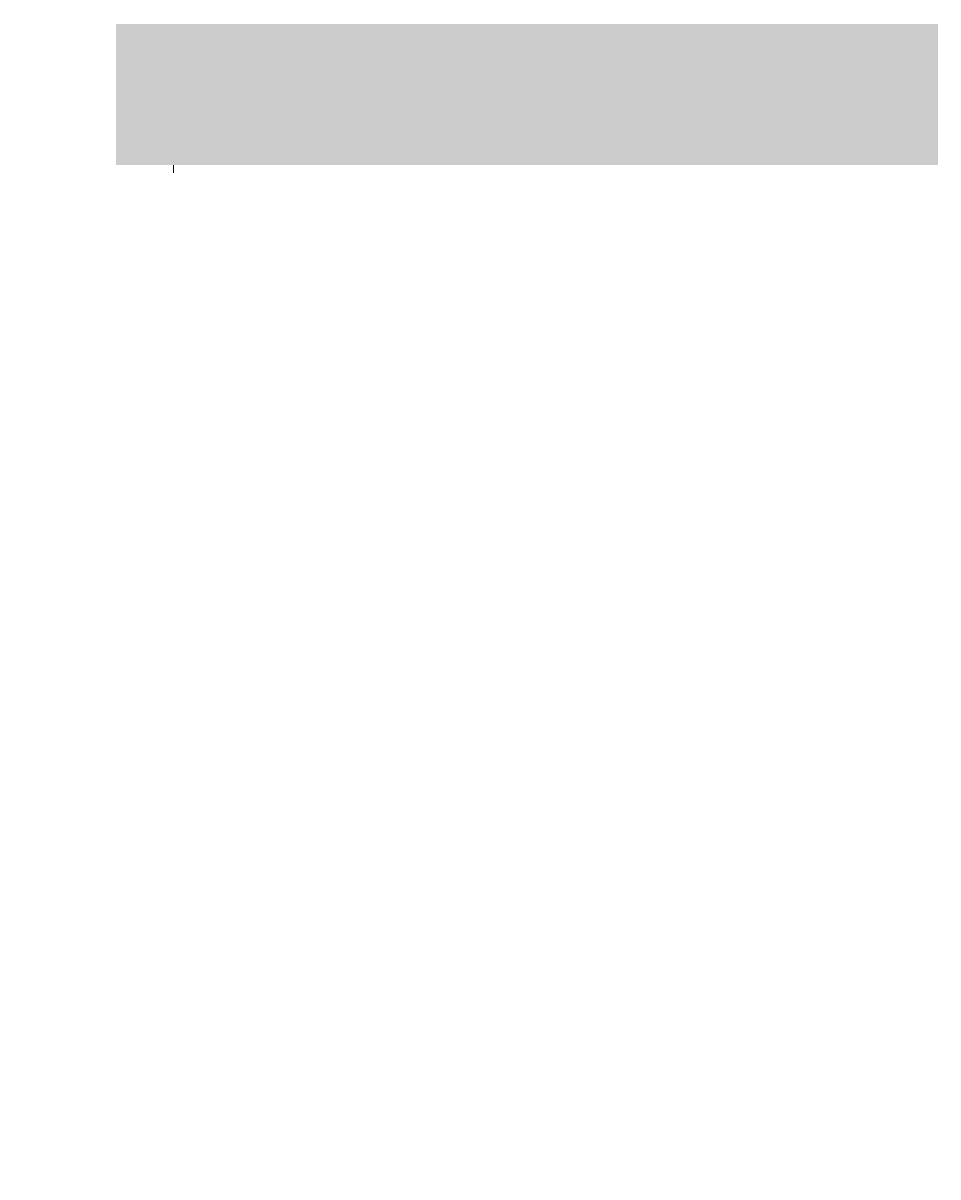}}%
    \put(0.1600138,1.03639045){\color[rgb]{0,0,0}\makebox(0,0)[lt]{\lineheight{1.25}\smash{\begin{tabular}[t]{l}220\end{tabular}}}}%
    \put(0,0){\includegraphics[width=\unitlength,page=2]{rl-trajectory.pdf}}%
    \put(0.28526339,1.03639045){\color[rgb]{0,0,0}\makebox(0,0)[lt]{\lineheight{1.25}\smash{\begin{tabular}[t]{l}240\end{tabular}}}}%
    \put(0,0){\includegraphics[width=\unitlength,page=3]{rl-trajectory.pdf}}%
    \put(0.41051297,1.03639045){\color[rgb]{0,0,0}\makebox(0,0)[lt]{\lineheight{1.25}\smash{\begin{tabular}[t]{l}260\end{tabular}}}}%
    \put(0,0){\includegraphics[width=\unitlength,page=4]{rl-trajectory.pdf}}%
    \put(0.53576257,1.03639045){\color[rgb]{0,0,0}\makebox(0,0)[lt]{\lineheight{1.25}\smash{\begin{tabular}[t]{l}280\end{tabular}}}}%
    \put(0,0){\includegraphics[width=\unitlength,page=5]{rl-trajectory.pdf}}%
    \put(0.66101215,1.03639045){\color[rgb]{0,0,0}\makebox(0,0)[lt]{\lineheight{1.25}\smash{\begin{tabular}[t]{l}300\end{tabular}}}}%
    \put(0,0){\includegraphics[width=\unitlength,page=6]{rl-trajectory.pdf}}%
    \put(0.78626175,1.03639045){\color[rgb]{0,0,0}\makebox(0,0)[lt]{\lineheight{1.25}\smash{\begin{tabular}[t]{l}320\end{tabular}}}}%
    \put(0,0){\includegraphics[width=\unitlength,page=7]{rl-trajectory.pdf}}%
    \put(0.91151133,1.03639045){\color[rgb]{0,0,0}\makebox(0,0)[lt]{\lineheight{1.25}\smash{\begin{tabular}[t]{l}340\end{tabular}}}}%
    \put(0.5160272,1.00241525){\color[rgb]{0,0,0}\makebox(0,0)[lt]{\lineheight{1.25}\smash{\begin{tabular}[t]{l}$x\,[\mathrm{m}]$\end{tabular}}}}%
    \put(0,0){\includegraphics[width=\unitlength,page=8]{rl-trajectory.pdf}}%
    \put(0.07548054,1.21055906){\color[rgb]{0,0,0}\makebox(0,0)[lt]{\lineheight{1.25}\smash{\begin{tabular}[t]{l}-2\end{tabular}}}}%
    \put(0,0){\includegraphics[width=\unitlength,page=9]{rl-trajectory.pdf}}%
    \put(0.08523769,1.16580596){\color[rgb]{0,0,0}\makebox(0,0)[lt]{\lineheight{1.25}\smash{\begin{tabular}[t]{l}2\end{tabular}}}}%
    \put(0,0){\includegraphics[width=\unitlength,page=10]{rl-trajectory.pdf}}%
    \put(0.08523769,1.12105286){\color[rgb]{0,0,0}\makebox(0,0)[lt]{\lineheight{1.25}\smash{\begin{tabular}[t]{l}6\end{tabular}}}}%
    \put(0,0){\includegraphics[width=\unitlength,page=11]{rl-trajectory.pdf}}%
    \put(0.07469217,1.07629976){\color[rgb]{0,0,0}\makebox(0,0)[lt]{\lineheight{1.25}\smash{\begin{tabular}[t]{l}10\end{tabular}}}}%
    \put(0,0){\includegraphics[width=\unitlength,page=12]{rl-trajectory.pdf}}%
    \put(0.15274047,0.80084359){\color[rgb]{0,0,0}\makebox(0,0)[lt]{\lineheight{1.25}\smash{\begin{tabular}[t]{l}0\end{tabular}}}}%
    \put(0,0){\includegraphics[width=\unitlength,page=13]{rl-trajectory.pdf}}%
    \put(0.30834458,0.80084359){\color[rgb]{0,0,0}\makebox(0,0)[lt]{\lineheight{1.25}\smash{\begin{tabular}[t]{l}1\end{tabular}}}}%
    \put(0,0){\includegraphics[width=\unitlength,page=14]{rl-trajectory.pdf}}%
    \put(0.46394871,0.80084359){\color[rgb]{0,0,0}\makebox(0,0)[lt]{\lineheight{1.25}\smash{\begin{tabular}[t]{l}2\end{tabular}}}}%
    \put(0,0){\includegraphics[width=\unitlength,page=15]{rl-trajectory.pdf}}%
    \put(0.61955282,0.80084359){\color[rgb]{0,0,0}\makebox(0,0)[lt]{\lineheight{1.25}\smash{\begin{tabular}[t]{l}3\end{tabular}}}}%
    \put(0,0){\includegraphics[width=\unitlength,page=16]{rl-trajectory.pdf}}%
    \put(0.77515694,0.80084359){\color[rgb]{0,0,0}\makebox(0,0)[lt]{\lineheight{1.25}\smash{\begin{tabular}[t]{l}4\end{tabular}}}}%
    \put(0,0){\includegraphics[width=\unitlength,page=17]{rl-trajectory.pdf}}%
    \put(0.93076105,0.80084359){\color[rgb]{0,0,0}\makebox(0,0)[lt]{\lineheight{1.25}\smash{\begin{tabular}[t]{l}5\end{tabular}}}}%
    \put(0,0){\includegraphics[width=\unitlength,page=18]{rl-trajectory.pdf}}%
    \put(0.08523769,0.82635765){\color[rgb]{0,0,0}\makebox(0,0)[lt]{\lineheight{1.25}\smash{\begin{tabular}[t]{l}0\end{tabular}}}}%
    \put(0,0){\includegraphics[width=\unitlength,page=19]{rl-trajectory.pdf}}%
    \put(0.07469217,0.87265396){\color[rgb]{0,0,0}\makebox(0,0)[lt]{\lineheight{1.25}\smash{\begin{tabular}[t]{l}10\end{tabular}}}}%
    \put(0,0){\includegraphics[width=\unitlength,page=20]{rl-trajectory.pdf}}%
    \put(0.07469217,0.91895027){\color[rgb]{0,0,0}\makebox(0,0)[lt]{\lineheight{1.25}\smash{\begin{tabular}[t]{l}20\end{tabular}}}}%
    \put(0,0){\includegraphics[width=\unitlength,page=21]{rl-trajectory.pdf}}%
    \put(0.07469217,0.96524657){\color[rgb]{0,0,0}\makebox(0,0)[lt]{\lineheight{1.25}\smash{\begin{tabular}[t]{l}30\end{tabular}}}}%
    \put(0,0){\includegraphics[width=\unitlength,page=22]{rl-trajectory.pdf}}%
    \put(0.15274047,0.55553111){\color[rgb]{0,0,0}\makebox(0,0)[lt]{\lineheight{1.25}\smash{\begin{tabular}[t]{l}0\end{tabular}}}}%
    \put(0,0){\includegraphics[width=\unitlength,page=23]{rl-trajectory.pdf}}%
    \put(0.30834458,0.55553111){\color[rgb]{0,0,0}\makebox(0,0)[lt]{\lineheight{1.25}\smash{\begin{tabular}[t]{l}1\end{tabular}}}}%
    \put(0,0){\includegraphics[width=\unitlength,page=24]{rl-trajectory.pdf}}%
    \put(0.46394871,0.55553111){\color[rgb]{0,0,0}\makebox(0,0)[lt]{\lineheight{1.25}\smash{\begin{tabular}[t]{l}2\end{tabular}}}}%
    \put(0,0){\includegraphics[width=\unitlength,page=25]{rl-trajectory.pdf}}%
    \put(0.61955282,0.55553111){\color[rgb]{0,0,0}\makebox(0,0)[lt]{\lineheight{1.25}\smash{\begin{tabular}[t]{l}3\end{tabular}}}}%
    \put(0,0){\includegraphics[width=\unitlength,page=26]{rl-trajectory.pdf}}%
    \put(0.77515694,0.55553111){\color[rgb]{0,0,0}\makebox(0,0)[lt]{\lineheight{1.25}\smash{\begin{tabular}[t]{l}4\end{tabular}}}}%
    \put(0,0){\includegraphics[width=\unitlength,page=27]{rl-trajectory.pdf}}%
    \put(0.93076105,0.55553111){\color[rgb]{0,0,0}\makebox(0,0)[lt]{\lineheight{1.25}\smash{\begin{tabular}[t]{l}5\end{tabular}}}}%
    \put(0,0){\includegraphics[width=\unitlength,page=28]{rl-trajectory.pdf}}%
    \put(0.06493503,0.71993409){\color[rgb]{0,0,0}\makebox(0,0)[lt]{\lineheight{1.25}\smash{\begin{tabular}[t]{l}-30\end{tabular}}}}%
    \put(0,0){\includegraphics[width=\unitlength,page=29]{rl-trajectory.pdf}}%
    \put(0.06493503,0.68636927){\color[rgb]{0,0,0}\makebox(0,0)[lt]{\lineheight{1.25}\smash{\begin{tabular}[t]{l}-15\end{tabular}}}}%
    \put(0,0){\includegraphics[width=\unitlength,page=30]{rl-trajectory.pdf}}%
    \put(0.08523769,0.65280445){\color[rgb]{0,0,0}\makebox(0,0)[lt]{\lineheight{1.25}\smash{\begin{tabular}[t]{l}0\end{tabular}}}}%
    \put(0,0){\includegraphics[width=\unitlength,page=31]{rl-trajectory.pdf}}%
    \put(0.07469217,0.61923961){\color[rgb]{0,0,0}\makebox(0,0)[lt]{\lineheight{1.25}\smash{\begin{tabular}[t]{l}15\end{tabular}}}}%
    \put(0,0){\includegraphics[width=\unitlength,page=32]{rl-trajectory.pdf}}%
    \put(0.07469217,0.58567479){\color[rgb]{0,0,0}\makebox(0,0)[lt]{\lineheight{1.25}\smash{\begin{tabular}[t]{l}30\end{tabular}}}}%
    \put(0,0){\includegraphics[width=\unitlength,page=33]{rl-trajectory.pdf}}%
    \put(0.15274047,0.31021862){\color[rgb]{0,0,0}\makebox(0,0)[lt]{\lineheight{1.25}\smash{\begin{tabular}[t]{l}0\end{tabular}}}}%
    \put(0,0){\includegraphics[width=\unitlength,page=34]{rl-trajectory.pdf}}%
    \put(0.31152018,0.31021862){\color[rgb]{0,0,0}\makebox(0,0)[lt]{\lineheight{1.25}\smash{\begin{tabular}[t]{l}1\end{tabular}}}}%
    \put(0,0){\includegraphics[width=\unitlength,page=35]{rl-trajectory.pdf}}%
    \put(0.47029988,0.31021862){\color[rgb]{0,0,0}\makebox(0,0)[lt]{\lineheight{1.25}\smash{\begin{tabular}[t]{l}2\end{tabular}}}}%
    \put(0,0){\includegraphics[width=\unitlength,page=36]{rl-trajectory.pdf}}%
    \put(0.6290796,0.31021862){\color[rgb]{0,0,0}\makebox(0,0)[lt]{\lineheight{1.25}\smash{\begin{tabular}[t]{l}3\end{tabular}}}}%
    \put(0,0){\includegraphics[width=\unitlength,page=37]{rl-trajectory.pdf}}%
    \put(0.7878593,0.31021862){\color[rgb]{0,0,0}\makebox(0,0)[lt]{\lineheight{1.25}\smash{\begin{tabular}[t]{l}4\end{tabular}}}}%
    \put(0,0){\includegraphics[width=\unitlength,page=38]{rl-trajectory.pdf}}%
    \put(0.94663902,0.31021862){\color[rgb]{0,0,0}\makebox(0,0)[lt]{\lineheight{1.25}\smash{\begin{tabular}[t]{l}5\end{tabular}}}}%
    \put(0,0){\includegraphics[width=\unitlength,page=39]{rl-trajectory.pdf}}%
    \put(0.07548054,0.3403623){\color[rgb]{0,0,0}\makebox(0,0)[lt]{\lineheight{1.25}\smash{\begin{tabular}[t]{l}-5\end{tabular}}}}%
    \put(0,0){\includegraphics[width=\unitlength,page=40]{rl-trajectory.pdf}}%
    \put(0.08523769,0.40749197){\color[rgb]{0,0,0}\makebox(0,0)[lt]{\lineheight{1.25}\smash{\begin{tabular}[t]{l}0\end{tabular}}}}%
    \put(0,0){\includegraphics[width=\unitlength,page=41]{rl-trajectory.pdf}}%
    \put(0.08523769,0.47462161){\color[rgb]{0,0,0}\makebox(0,0)[lt]{\lineheight{1.25}\smash{\begin{tabular}[t]{l}5\end{tabular}}}}%
    \put(0,0){\includegraphics[width=\unitlength,page=42]{rl-trajectory.pdf}}%
    \put(0.15274047,0.06490614){\color[rgb]{0,0,0}\makebox(0,0)[lt]{\lineheight{1.25}\smash{\begin{tabular}[t]{l}0\end{tabular}}}}%
    \put(0,0){\includegraphics[width=\unitlength,page=43]{rl-trajectory.pdf}}%
    \put(0.31152018,0.06490614){\color[rgb]{0,0,0}\makebox(0,0)[lt]{\lineheight{1.25}\smash{\begin{tabular}[t]{l}1\end{tabular}}}}%
    \put(0,0){\includegraphics[width=\unitlength,page=44]{rl-trajectory.pdf}}%
    \put(0.47029988,0.06490614){\color[rgb]{0,0,0}\makebox(0,0)[lt]{\lineheight{1.25}\smash{\begin{tabular}[t]{l}2\end{tabular}}}}%
    \put(0,0){\includegraphics[width=\unitlength,page=45]{rl-trajectory.pdf}}%
    \put(0.6290796,0.06490614){\color[rgb]{0,0,0}\makebox(0,0)[lt]{\lineheight{1.25}\smash{\begin{tabular}[t]{l}3\end{tabular}}}}%
    \put(0,0){\includegraphics[width=\unitlength,page=46]{rl-trajectory.pdf}}%
    \put(0.7878593,0.06490614){\color[rgb]{0,0,0}\makebox(0,0)[lt]{\lineheight{1.25}\smash{\begin{tabular}[t]{l}4\end{tabular}}}}%
    \put(0,0){\includegraphics[width=\unitlength,page=47]{rl-trajectory.pdf}}%
    \put(0.94663902,0.06490614){\color[rgb]{0,0,0}\makebox(0,0)[lt]{\lineheight{1.25}\smash{\begin{tabular}[t]{l}5\end{tabular}}}}%
    \put(0,0){\includegraphics[width=\unitlength,page=48]{rl-trajectory.pdf}}%
    \put(0.06167982,0.22930912){\color[rgb]{0,0,0}\makebox(0,0)[lt]{\lineheight{1.25}\smash{\begin{tabular}[t]{l}-10\end{tabular}}}}%
    \put(0,0){\includegraphics[width=\unitlength,page=49]{rl-trajectory.pdf}}%
    \put(0.07873575,0.1957443){\color[rgb]{0,0,0}\makebox(0,0)[lt]{\lineheight{1.25}\smash{\begin{tabular}[t]{l}-5\end{tabular}}}}%
    \put(0,0){\includegraphics[width=\unitlength,page=50]{rl-trajectory.pdf}}%
    \put(0.08523769,0.16217947){\color[rgb]{0,0,0}\makebox(0,0)[lt]{\lineheight{1.25}\smash{\begin{tabular}[t]{l}0\end{tabular}}}}%
    \put(0,0){\includegraphics[width=\unitlength,page=51]{rl-trajectory.pdf}}%
    \put(0.08523769,0.12861465){\color[rgb]{0,0,0}\makebox(0,0)[lt]{\lineheight{1.25}\smash{\begin{tabular}[t]{l}5\end{tabular}}}}%
    \put(0,0){\includegraphics[width=\unitlength,page=52]{rl-trajectory.pdf}}%
    \put(0.07469217,0.09504982){\color[rgb]{0,0,0}\makebox(0,0)[lt]{\lineheight{1.25}\smash{\begin{tabular}[t]{l}10\end{tabular}}}}%
    \put(0,0){\includegraphics[width=\unitlength,page=53]{rl-trajectory.pdf}}%
    \put(0.54082917,0.03007845){\color[rgb]{0,0,0}\makebox(0,0)[lt]{\lineheight{1.25}\smash{\begin{tabular}[t]{l}$t\,[\mathrm{s}]$\end{tabular}}}}%
    \put(0.04425978,0.1405568){\color[rgb]{0,0,0}\rotatebox{90}{\makebox(0,0)[lt]{\lineheight{1.25}\smash{\begin{tabular}[t]{l}$\delta\,[^\circ]$\end{tabular}}}}}%
    \put(0.04880578,0.3751977){\color[rgb]{0,0,0}\rotatebox{90}{\makebox(0,0)[lt]{\lineheight{1.25}\smash{\begin{tabular}[t]{l}$a\,[\frac{\mathrm{m}}{\mathrm{s}^2}]$\end{tabular}}}}}%
    \put(0.0491672,0.63428497){\color[rgb]{0,0,0}\rotatebox{90}{\makebox(0,0)[lt]{\lineheight{1.25}\smash{\begin{tabular}[t]{l}$\varphi\,[^\circ]$\end{tabular}}}}}%
    \put(0.05138337,0.86897354){\color[rgb]{0,0,0}\rotatebox{90}{\makebox(0,0)[lt]{\lineheight{1.25}\smash{\begin{tabular}[t]{l}$v\,[\frac{\mathrm{m}}{\mathrm{s}}]$\end{tabular}}}}}%
    \put(0.05091476,1.11986113){\color[rgb]{0,0,0}\rotatebox{90}{\makebox(0,0)[lt]{\lineheight{1.25}\smash{\begin{tabular}[t]{l}$y\,[\mathrm{m}]$\end{tabular}}}}}%
  \end{picture}%
\endgroup%
};
        \node[sedan top,body color=red1,window color=black!10,minimum width=0.6cm,rotate=-9] at (-2.8,4.5) {};
        \node[sedan top,body color=blue1,window color=black!10,minimum width=0.6cm,opacity=0.5] at (-0.5,4.15) {};
        \node[sedan top,body color=blue1,window color=black!10,minimum width=0.6cm,opacity=0.5] at (3.3,4.55) {};
    \end{tikzpicture}
    \caption{SRL trajectory (dashed lines) used for linearization within the LTV-MPC and the locally optimal solution (continuous lines) of the MPC based on the SRL trajectory.}
    \label{fig:rl-trajectory}
\end{figure}

\cref{fig:rl-trajectory} shows an example of an RL trajectory based solution calculated by the LTV-MPC.
The figure illustrates the state trajectories as well as the control trajectories.

\section{Results and Evaluation}
\label{sec:results_and_evaluation}

\begin{figure*}[t]
    \centering\vspace*{2mm}
    \includegraphics[width=0.9\linewidth]{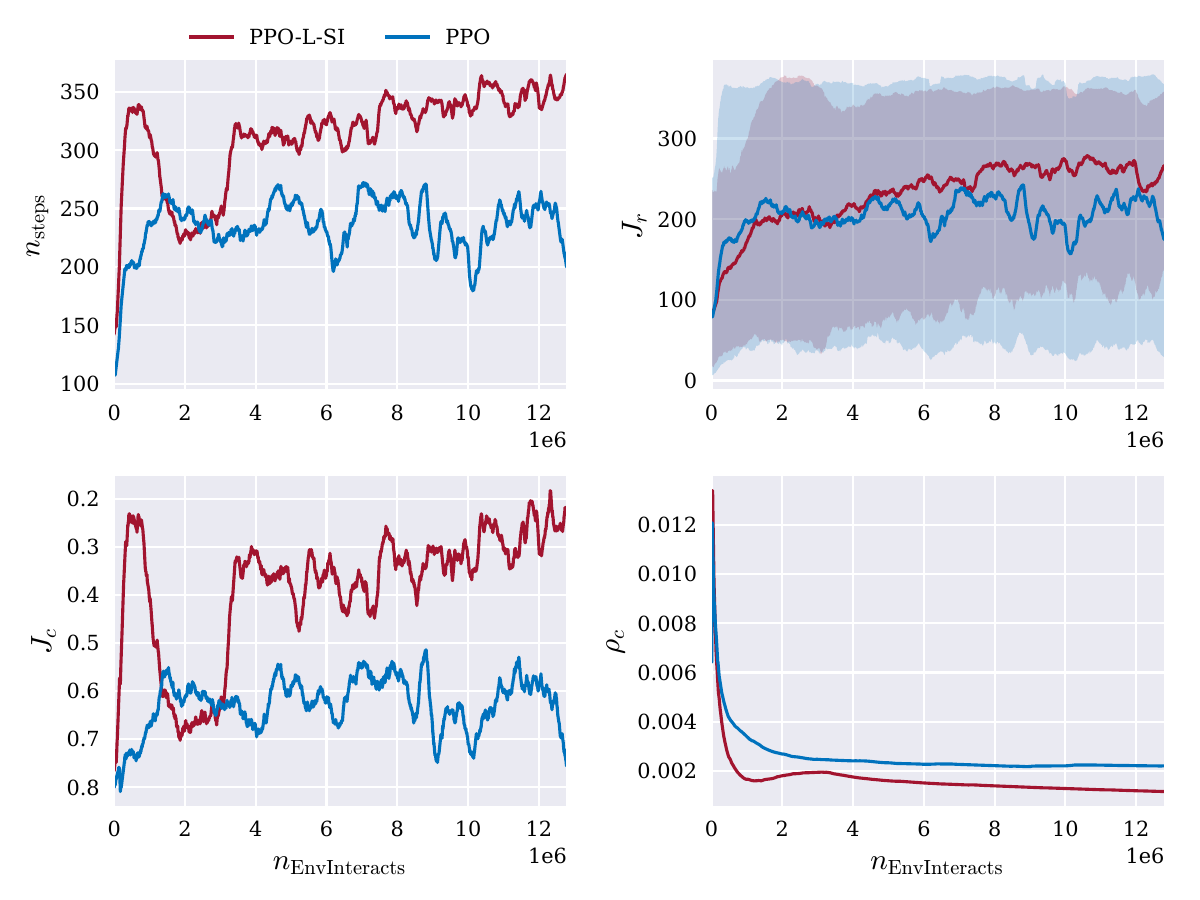}
    \caption{Training results for the baseline RL algorithm PPO and the SRL algorithm PPO-L-SI.}
    \label{fig:ppo_vs_ppolsi-training}
\end{figure*}

\subsection{Experimental Set-Up}
For our evaluations we use the \textit{highway-env} \cite{highway-env} framework, which provides environments for tactical decision-making in different automated driving tasks.
Within this framework, the agent controls the ego vehicle, while the other vehicles follow the Intelligent Driver Model (IDM) and only react to the ego vehicle once it enters their lane.
We focus on a three lane highway driving scenario as illustrated in \cref{fig:observation}.
At the start of each episode, the agent and the other traffic participants are generated randomly on one of the lanes with a random velocity within a defined range.
The lane changes of the IDM vehicles are disabled.
Furthermore, we initialize the MPC controller with a trajectory with constant velocity.

In both, MPC and RL, the middle of the rightmost lane is defined as optimal by means of the objective function and the reward function, respectively ($y^\mathrm{ref}=\mathrm{middle\ of\ right\ lane}$).
Therefore, the agent will tend to drive on the right lane.
Furthermore, the longitudinal velocity goal is defined as $v_x^\mathrm{ref}=25\si[]{\metre\per\second}$.

\subsection{Metrics}\label{subsec:metrics}
To evaluate and compare the described control methods with the proposed SRMPC, this section outlines the metrics.
Regarding safety and performance, the following metrics are considered:

\begin{itemize}
    \item Episodic return $J_r$: This metric uses the unprocessed reward function $r'$ \eqref{reward-function} to describe the performance of the agents.
    \item Episodic cost return $J_c$: For better comparability this metric considers only actual collisions and road exits as a cost ($J_c=1$).
          This metric can only assume the values $J_c\in\{0,1\}$, since the episode terminates when costs occur.
    \item Cost rate $\rho_c$: This metric describes the relationship between the total accumulated costs and the accumulated number of environment steps over all simulations.
    \item Episodic steps $n_\mathrm{steps}$: The metric describes the number of steps taken by the agent without the occurrence of costs.
          $n_\mathrm{steps}$ can take on values in the interval $n_\mathrm{steps}\in[0,400]$ due to the maximum episode length.
    \item Average longitudinal velocity $\overline{v}_x$: A performance measure for comparison with the longitudinal reference velocity $v_x^\mathrm{ref}$.
\end{itemize}
The episodic return $J_r$ and average velocity $\overline{v}_x$ are referred to as performance metrics and the episodic cost return $J_c$, the cost rate $\rho_c$ and the episodic steps $n_\mathrm{steps}$ are referred to as safety metrics.

\subsection{Results}
First we evaluate the PPO-L-SI against the baseline PPO algorithm in training to demonstrate the effectiveness regarding safety.
The training is shown in \cref{fig:ppo_vs_ppolsi-training}.
Here, the values of the cost rate $J_c$ are modified by a running mean filter to see a clearer trend of collisions respectively road exits.
The filtered metric can be interpreted as the failure rate of the agent.
It can be seen that the safety metrics of the PPO-L-SI take better values than those of the PPO algorithm almost over the entire training.
Especially at the end of training the cost return $J_c$ and cost rate $\rho_c$ from PPO-L-SI are more than twice as good as those from PPO.
Also, the performance metric episodic return $J_r$ of the PPO-L-SI lies above that of PPO almost during the entire training.
Accordingly, the PPO-L-SI achieves much better results as the baseline PPO algorithm.
In particular when looking at the safety metrics.

In the next step we evaluate the learning based PPO-L-SI and the optimization based LTV-MPC against the combined approach SRMPC.
Therefore, we simulate all algorithms over 1000 episodes and track all described metrics from section \ref{subsec:metrics}.
In contrast to the RL training evaluation (see \cref{fig:ppo_vs_ppolsi-training}) we now use mean values for the comparison.
The cost rate $\rho_c$ already is a metric tracking a running mean.
The metrics for each control method are reported from table \ref{tab:evaluation}.
All three methods are compared in a scenario with light and dense traffic volume.
In light traffic the vehicles are randomly generated in one of the three lanes at a distance of around $25\si{\meter}$ from each other whereas in dense traffic the distance reduces to roughly $15\si{\meter}$.

\begin{table}[h]
    \vspace*{2mm}
    \caption{Evaluation results: Comparison between the SRL algorithm PPO-L-SI, LTV-MPC and SRMPC based on a highway driving scenario with light and dense traffic.}
    \centering
    \footnotesize
    \begin{tabular}{clrrrrr}
        \hline\noalign{\vskip 1mm}
         & \textbf{Algorithm} & $\overline{J}_r$ (per step)       & $\overline{J}_c$           & $\rho_c$                    & $\overline{n}_{\mathrm{steps}}$ & $\overline{v}_x$           \\[1mm]
        \hline\noalign{\vskip 1mm}
        \parbox[t]{2mm}{\multirow{3}{*}{\rotatebox[origin=c]{90}{light}}}
         & PPO-L-SI           & -4669 (-13.4)                     & 0.308                      & 0.0009                      & 348                             & 23.11                      \\[1mm]
         & LTV-MPC            & -1241 (-5.1)                      & 0.625                      & 0.0026                      & 244                             & 24.38                      \\[1mm]
         & SRMPC              & \cellcolor{green1!25}-1108 (-3.1) & \cellcolor{green1!25}0.215 & \cellcolor{green1!25}0.0006 & \cellcolor{green1!25}357        & \cellcolor{green1!25}24.63 \\[1mm]
        \hline
        \hline\noalign{\vskip 1mm}
        \parbox[t]{2mm}{\multirow{3}{*}{\rotatebox[origin=c]{90}{dense}}}
         & PPO-L-SI           & -4499 (-16.4)                     & 0.540                      & 0.002                       & 274                             & 22.47                      \\[1mm]
         & LTV-MPC            & \cellcolor{green1!25}-334 (-2.4)  & 0.788                      & 0.0057                      & 139                             & \cellcolor{green1!25}24.79 \\[1mm]
         & SRMPC              & -1211 (-4.0)                      & \cellcolor{green1!25}0.353 & \cellcolor{green1!25}0.0012 & \cellcolor{green1!25}305        & 24.47                      \\[1mm]
        \hline
    \end{tabular}\label{tab:evaluation}
\end{table}

When evaluating the light traffic case, the SRMPC outperforms both methods, the PPO-L-SI and the LTV-MPC in terms of safety and performance.
The proposed SRMPC reaches a higher return and a higher average velocity compared to PPO-L-SI and LTV-MPC.
Regarding the cost return and cost rate, the SRMPC achieves more than $30\si[]{\percent}$ better results compared to PPO-L-SI and more than $65\si[]{\percent}$ better results than LTV-MPC.

The evaluation results at dense traffic differ a little from those at light traffic.
Nevertheless, the SRMPC outperforms the two methods on which it is based in terms of safety.
But the LTV-MPC achieves slightly better results in terms of performance.
However, the better results are not decisive so that the performance with respect to the return per step and average velocity is comparable to that of the SRMPC.

\subsection{Computation Time}

The computation time of the SRMPC increases compared to the LTV-MPC due to the calculation of the reference trajectory using the SRL algorithm.
This step requires the evaluation of the policy and the forward simulation of the environment by the chosen horizon length.
Since the used simulation environment was not optimized for simulation performance, the computation time of the forward simulation is relatively high.
Hence, we did not run quantitative experiments to compare the computation time of the SRMPC with the LTV-MPC.
To achieve real-time capability, we propose using a small policy network architecture and an efficient prediction framework for the forward simulation.

\section{Conclusions and Future Work}
\label{sec:conclusion}
We have demonstrated the effective use of RL to enhance the local optimization of LTV-MPC. 
The resulting control method outperforms the underlying RL and LTV-MPC methods in terms of safety and performance for the highway driving task in light traffic.
In dense traffic, the developed method achieves superior results with regard to safety and shows roughly similar results in the performance compared to the LTV-MPC. 

The utilization of an energy-based function for expressing and learning safety has shown promising results, suggesting that further research should be done on the CRL approach with SI and a state dependent Lagrangian multiplier. 
However, a possible reformulation of the SI is worth considering, aligning it with the mathematical formulation of e.g. Responsibility-Sensitive Safety (RSS) \cite*{RSS.2017} for autonomous vehicles.
Furthermore, it is conceivable to extend the proposed method to NMPC by utilizing an RL trajectory as the initial solution of an NLP solver to shorten the computation time. %

\balance
\printbibliography

\clearpage

\end{document}